\documentclass[runningheads]{llncs}
\usepackage[T1]{fontenc}
\usepackage{graphicx}
\usepackage{wrapfig}
\usepackage{hyperref}
\usepackage{url}
\usepackage{comment}
\usepackage{amsmath}
\usepackage{booktabs}
\usepackage{enumitem}

\begin{document}

\title{CLEF HIPE-2026: Evaluating Accurate and Efficient Person--Place Relation Extraction from Multilingual Historical Texts\thanks{\small The version of record is available at \href{https://doi.org/10.1007/978-3-032-21321-1_46}{https://doi.org/10.1007/978-3-030-99739-7\_44}.}}

\author{Juri Opitz\inst{1} \and
Corina Raclé\inst{1} \and
Emanuela Boros\inst{3} \and
Andrianos Michail\inst{1} \and
 \\ Matteo Romanello\inst{2} \and
Maud Ehrmann\inst{3} \and
Simon Clematide\inst{1}
}

\authorrunning{Opitz et al.}
\titlerunning{CLEF HIPE-2026: Evaluating Person-Place Relation}

\institute{
University of Zurich, Switzerland \and
Swiss Art Research Infrastructure (SARI), University of Zurich \and
École Polytechnique Fédérale de Lausanne (EPFL), Switzerland \\
\email{hipe-2026@googlegroups.com} \\
\url{https://hipe-eval.github.io/HIPE-2026}
}

\maketitle

\begin{abstract}

HIPE-2026 is a CLEF evaluation lab dedicated to person-place relation extraction from noisy, multilingual historical texts. Building on the HIPE-2020 and HIPE-2022 campaigns, it extends the series toward semantic relation extraction by targeting the task of identifying person–place associations in multiple languages and time periods. Systems are asked to classify relations of two types—\textbf{at} (“Has the person ever been at this place?”) and \textbf{isAt} (“Is the person located at this place around publication time?”)—requiring reasoning over temporal and geographical cues. The lab introduces a three-fold evaluation profile that jointly assesses accuracy, computational efficiency, and domain generalization. 
By linking relation extraction to large-scale historical data processing, HIPE-2026 aims to support downstream applications in knowledge-graph construction, historical biography reconstruction, and spatial analysis in digital humanities.
\keywords{Relation Extraction \and Multilingual NLP \and Digital Humanities \and Historical Texts \and Shared Task}
\end{abstract}

\section{Introduction}


Historical documents bear a wealth of information that can help reconstructing past events, biographies, and the development of social networks. However, the digitization of historical documents typically results in data that are noisy, multilingual, and weakly structured. Therefore, robust approaches to mining historical data are necessary to support a wide range of information needs among scholars in the social sciences and history \cite{fokkens2014biographynet,schich2014historical,opitz2019automatic,lucchini2019mobility,cardoso2020construction,tamper2023biographysampo,zhong2023comprehensive}. 

HIPE-2026 is a CLEF Evaluation Lab dedicated to the extraction of person–place relations in multilingual historical documents. A person-place relation corresponds to a semantic link between an individual and a location as evidenced in a document. Such relations may indicate where a person is said to be at a given moment, where they lived or worked, or places connected to notable moments in their life (e.g., birthplaces, residences, visits, travel destinations). Together, these relations can help answer the question of \textit{Who was where when?}, and support the reconstruction of individuals' geographical and temporal trajectories.

These implicit or explicit, spatio-temporal relations cannot be detected through simple document co-occurrence of entity mentions. Rather, it requires temporal reasoning, geographical inference, and interpretation of noisy historical texts--often with sparse or indirect contextual cues--to detect and qualify person–place relations with appropriate degrees of certainty.

The objective of HIPE-2026 is to advance the automatic detection of such relations, enabling the reconstruction of individuals' movements in space and time and the tracing of life trajectories in support of digital humanities scholarship. The task is designed to be approachable by both generative AI systems (LLMs) and more traditional classification models. HIPE-2026 builds on HIPE-2020 and HIPE-2022, which targeted named-entity recognition and linking in historical corpora~\cite{ehrmann2020introducing,ehrmann2022introducing} and advances the series toward relation extraction (RE).

\section{Task Description}

\begin{figure}[t]\centering
    \includegraphics[width=0.65\linewidth]{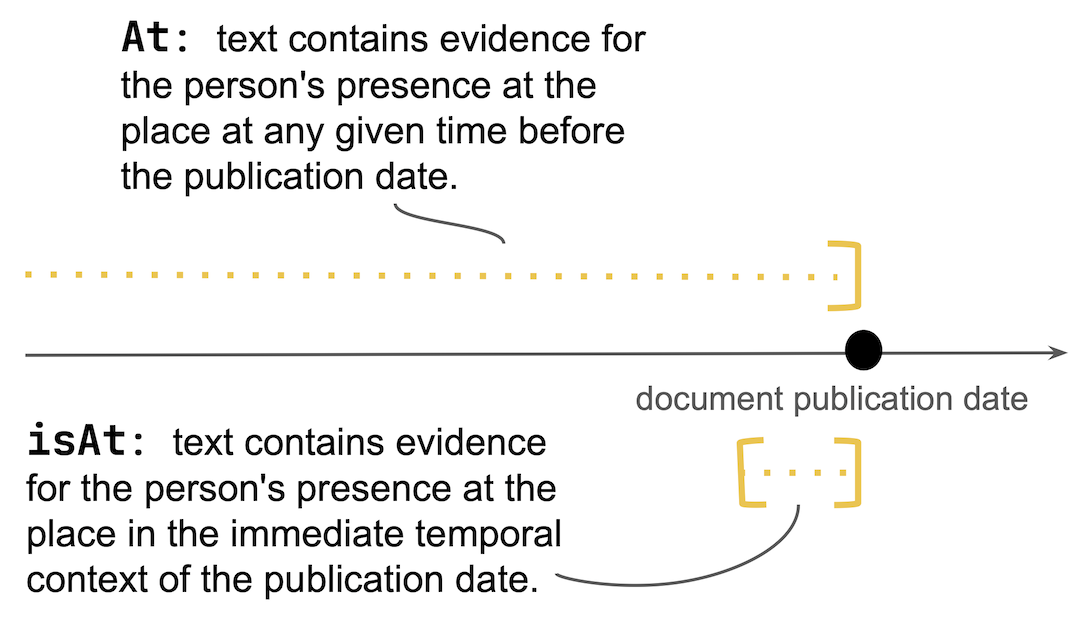} 
   \caption{Temporal scope of \textbf{at} and \textbf{isAt} relations relative to publication time.}
    \label{fig:wrapfig}
\end{figure}

Participants are tasked with determining the relationship of each person--place pair mentioned in a historical document. Each pair consists
of two entities, one person and one location, each of which appears in the text through one
or more mentions. For each pair, systems must assess whether the text provides evidence of the person’s presence at the location, taking into account the document’s temporal horizon relative to its publication date (see Figure~\ref{fig:wrapfig} for an illustration).

For each document, systems must classify each candidate pair according to the following relation types:

\begin{description}[itemsep=0em,topsep=3pt]
  \item[\textbf{at}] The text provides evidence that the person was present at the place at any point in time \emph{prior to the publication date}. This relation is right-bounded by the publication date but may hold at any time in the past. It is labeled as:
  \begin{itemize}[itemsep=0em,topsep=2pt]
    \item \textbf{true}: explicit evidence supports the relation;
    \item \textbf{probable}: the relation can be inferred from contextual clues and is a likely assumption;
    \item \textbf{false}: no evidence, or contradictory evidence, is present.
  \end{itemize}
  \item[\textbf{isAt}] captures whether the text provides implicit or explicit evidence that the person was at the location in the \emph{immediate temporal context of the publication date}. The decision is binary (\textbf{+}/\textbf{--}): \textbf{+} means that there is implicit or explicit evidence the person was at the location shortly before the publication date, while \textbf{--} indicates the absence of such evidence.
\end{description}

\paragraph{Relation between \texttt{at} and \texttt{isAt}.}

The \textbf{isAt} relation can be viewed as a temporal refinement of \textbf{at}: it captures whether there is evidence (implicit or explicit) for the \textbf{at} relation in the closer temporal scope of the article's date. Conceptually, assigning \textbf{isAt}=\textbf{+} presupposes that an \textbf{at} relation is true, or at least, probable; Predictions where \textbf{at}=\textbf{false} and \textbf{isAt}=\textbf{+} are therefore epistemically inconsistent, but practically permitted. In sum, if \textbf{at} is \textbf{true} or \textbf{probable}, \textbf{isAt} would further specify whether the person’s presence falls within the document’s temporal horizon.

\paragraph{Abductive interpretation and evidential reasoning.}

The distinction between explicit, probable, and absent evidence is grounded in an abductive view of language understanding. Hobbs' framework of ``Interpretation as Abduction'' \cite{hobbs1993abduction} argues that discourse interpretation consists in identifying the minimal set of assumptions which, together with background knowledge, explain why an utterance would be a rational and coherent thing to say. Under this view, meaning is not limited to what is explicitly stated, but includes what must be assumed to provide a coherent explanation of the discourse. Applied to person–place relations in historical texts, this perspective motivates the \textbf{probable} label: relations may be supported by indirect cues such as event participation, institutional roles, or narrative coherence, even when no explicit locative statement is present. Conversely, the \textbf{false} label reflects cases where no such abductive explanation is warranted based on the text alone. The \textbf{isAt} refinement further constrains abductive inference temporally, requiring that assumptions about presence remain compatible with the document’s publication horizon.
 
In addition, we allow systems to optionally provide free-text explanations or indicate other relevant background knowledge that informed their decisions. Evaluating such explanations is not a primary objective of the task, but participants are encouraged to analyze and report pieces of information that could help us better understand prediction rationales.

\subsection*{Example-based Illustration of the Task}

To illustrate the task, we present an excerpt from an article of the American newspaper \textit{Tabor City Tribune} on June 6, 1960. The article reports on a military scout meeting and involves multiple person-place candidate relation  pairs, as shown in Table \ref{tab:example_annotation}.

\begin{quote}\scriptsize
Loris Scouts Win Camporee Top Honor WINNERS — Members of Boy Scout Troop 843. »»ho won top honors at an Horry IMstrict Scout Camporee at Clear Pond March 25-27 are shown above hiking down a road. The lads won after engaging in com- passing, rope throwing, Morse code, hiking with compass and individual cooking. The troop is sponsored by the First Bap tist Church. Boy Scout Troop 843 of Loris won top honors at the Horry District Scout Camporee March 25-27 at Clear Pond between ( Conway and Myrtle Beach. Competing against some ot the finest units in the county, the boys won top honors after participating in Compassing, : rope throwing, Morse Code, hiking with compass and indi- ) vidual cooking. Friday night the troops! competing joined in a camp fire meeting and heard an ad dress by Col. Gruenwald, com manding officer of the Myrtle Beach Air Force Base. The Loris troop, sponsored by the First Baptist church, was the only troop to have all its Scout leaders present: Francis Ragan, George Rent/ and George Lav.
\end{quote}

A clear case is the relation between the person entity \textit{Col. Gruenwald} and the place entity \textit{Myrtle Beach Air Force Base}. The article explicitly identifies the colonel as the ``commanding officer of the'' air force base, thereby supporting a true label for the \textbf{at} relation, and a positive one for \textbf{isAt}. In contrast, although the place \textit{Myrtle Beach} is also mentioned, the text does not state that Col. Gruenwald was physically present in the city; his connection is institutional rather than locational. This motivates the probable label for the \textbf{at} relation and a negative one for \textbf{isAt}, highlighting the distinction between affiliation and physical presence.

The relation between Col. Gruenwald and Clear Pond (as well as the Horry District Scout Camporee) provides another instructive example. The article reports that he delivered an address at a campfire meeting during the camporee, which took place at Clear Pond. This constitutes sufficient evidence to annotate both locations with a \textsc{true} \texttt{at} relation and a positive \textbf{isAt} relation, even though his presence is inferred indirectly through the event.

Several negative annotations demonstrate the importance of avoiding inference beyond the text. Conway and Loris are mentioned as geographic reference points, but there is no indication that Col. Gruenwald visited either place. Consequently, these relations are marked as false, despite their proximity to the actual event location.

\begin{table}[h]
\centering
\scriptsize
\caption{Annotated person-place candidate relations for the example article.}
\begin{tabular}{llcc}
\toprule
\textbf{Person} & \textbf{Place} & \textbf{at} & \textbf{isAt} \\ 
\midrule
Col. Gruenwald	 & Myrtle Beach Air Force Base & true & +\\
Col. Gruenwald	 & Clear Pond& true & +\\
Col. Gruenwald	 & Myrtle Beach & probable & -- \\
Col. Gruenwald	 & Loris & false & --\\
Col. Gruenwald	 & Conway& false & --\\
Col. Gruenwald	 & Horry District Scout Camporee & true & +\\
Francis Ragan	 & Myrtle Beach Air Force Base & false & --\\
Francis Ragan	 & Conway & false & --\\
Francis Ragan	 & Myrtle Beach & false & --\\
George Lav	 & Myrtle Beach Air Force Base & false & --\\
George Lav	 & Conway & false & --\\
George Rent	 & Myrtle Beach Air Force Base & false & --\\
George Rent	 & Myrtle Beach & false & --\\
George Rent	 & Loris & true & --\\
George Rent	 & Conway & false & --\\
George Rent	 & Horry District Scout Camporee & true & +\\
\bottomrule
\end{tabular}
\label{tab:example_annotation}
\end{table}

\section{Data}

HIPE-2026 data consists of two sets of data:

\begin{description}[itemsep=0em,topsep=3pt]
  \item \textbf{Test Set A}: historical newspaper articles in French, German, English, and Luxembourgish spanning roughly 200 years (19th–20th centuries), drawn from the HIPE-2022 data.
  \item \textbf{Surprise Test Set B}: French literary texts (16th–18th century), used to assess domain generalization; only the \textbf{at} relation is evaluated.
\end{description}

\paragraph{Pilot Study.}
A total of 119 person–place pairs were annotated by three independent annotators, drawn from the English and French development sets of HIPE-2022. The inter-annotator agreement (Cohen’s kappa) ranged from 0.7 to 0.9 for the \textbf{at} relation, and from 0.4 to 0.9 for \textbf{isAt}, indicating moderate to high consistency. Large language models also showed promising alignment with human judgments: GPT-4o reached up to 0.8 agreement with the gold standard for \textbf{at}, while \textbf{isAt} results were lower and more variable (0.2–0.7). The study further highlighted the high inference cost of current models and the need for scalable methods that can handle the multiplicative growth of candidate entity pairs in historical documents. 

\section{Evaluation}

We consider three evaluation profiles.

The \textbf{accuracy profile} rewards high-performing systems and encourages the exploration of frontier models, advanced prompting strategies, or agent-based approaches. System responses are evaluated per test set using \emph{macro-averaged Recall} (also known as \emph{balanced accuracy}).  
Macro Recall is defined as follows:
\begin{enumerate}
    \item For each label $\ell$ in the label set $L$, compute its recall:
    \[
        \mathrm{Recall}(\ell) = 
        \frac{\#\text{examples with label }\ell\ \text{correctly predicted}}
             {\#\text{examples whose gold label is }\ell}.
    \]
    \item The final score is the arithmetic mean of the per-label recalls:
    \[
        \mathrm{MacroRecall} = \frac{1}{|L|}\sum_{\ell \in L} \mathrm{Recall}(\ell).
    \]
\end{enumerate}

This metric is principled and interpretable, ensuring that all labels contribute equally regardless of class imbalance \cite{10.1162/tacl_a_00675,sebastiani2015axiomatically}.  
For Test Set A, which includes three labels for \textbf{at} and two labels for \textbf{isAt}, macro Recall is computed separately for each relation and then averaged to obtain the final system ranking.

The \textbf{accuracy-efficiency profile} promotes lightweight, scalable methods, including smaller LLMs or task-specific classifiers. This profile balances predictive performance with  efficiency and resource usage. This is motivated by the increasing cost of running very large models and, in our case, by the scale of digitized material together with the quadratic nature of the task (person–location \textit{pairs}). Participant teams will be surveyed regarding parameter count and model size; together with accuracy, these factors are integrated into a robust ranking metric that computes a balanced score reflecting system efficiency. 

Finally, the \textbf{generalization profile} tests system accuracy on the Surprise Test Set B, based on the \textbf{at}-relation only and using \emph{macro-Recall}.

\vspace{0.2cm}

\noindent Annotated data, baselines and scoring tools are released via GitHub\footnote{\url{https://github.com/hipe-eval/HIPE-2026-data}} under a CC-BY 4.0 license, and participation guidelines are published on Zenodo\footnote{\url{https://zenodo.org/records/17800136}}.

\section{Related Work}
\paragraph{Relation Extraction (RE) Benchmarks and Their Limitations.}

Early Open Information Extraction (Open IE) approaches extracted unrestricted relations without predefined schemas \cite{etzioni2008open}. Subsequent benchmarks introduced controlled settings with fixed relation inventories, notably TACRED \cite{ZhangZhong:2017tacred} for sentence-level and DocRED \cite{yao2019docred} for document-level RE. These English-only datasets, covering several dozen relation types, have advanced the field but suffer from annotation incompleteness. Re-annotation efforts such as Re-TACRED \cite{alt-etal-2020-tacred,stoica2021retacred} and Re-DocRED \cite{tan-etal-2022-revisiting} reduce false negatives, yet the benchmarks remain confined to clean, modern English. In contrast, HIPE-2026 focuses on a single relation type—person–place—across multiple languages, time periods, and OCR-derived historical sources that involve orthographic variation and temporal reasoning. Previous HIPE tasks (2020, 2022) addressed multilingual named entity recognition and linking in historical newspapers \cite{ehrmann_overview_2020,ehrmann_overview_2022} but not relation extraction. A recent survey highlights progress in multilingual RE through cross-lingual transfer and annotation projection while underscoring the lack of benchmarks addressing multilinguality, noise, and domain shift \cite{AliSpeck:2025}. HIPE-2026 fills this gap with a historically and linguistically diverse evaluation setting.

\paragraph{Relation Extraction for Biographical and Historical Domains.}
Biographical RE extends general RE toward structured knowledge bases. The \textit{Biographical} dataset aligns Wikipedia with Pantheon and Wikidata via distant supervision to derive ten relations~\cite{plum2022biographical}. OpenIE has been used to extract RDF triples from Wikipedia biographies \cite{sugimoto2023closer}. The \textit{Guided Distant Supervision (GDS)} variant adapts this approach to German, denoises labels through external constraints, and explores cross-lingual transfer~\cite{plum2024guided}. Although these studies demonstrate the potential of distant supervision for domain-specific RE, historical data remain largely unexplored, a gap that HIPE-2026 aims to close.

\paragraph{Multilingual, Noisy, and Domain-Shift Relation Extraction.}

Multilingual RE methods rely on cross-lingual transfer, annotation projection, or zero-shot learning (see \cite{AliSpeck:2025,zhong2023comprehensive}). However, few benchmarks incorporate noise robustness, domain shift, or low-resource historical languages, with emerging datasets addressing explicitly historical materials \cite{yang-etal-2023-histred,rodriguez2022ocrre,quaresma2020information}. This motivates shared tasks like HIPE-2026, which emphasize relation mining in noisy, multilingual, historical contexts.

\paragraph{Denoising Techniques.}
Distant supervision enables large-scale RE but introduces substantial noise. Recent denoising methods include \textit{hierarchical contrastive learning} (HiCLRE) for filtering noisy instances~\cite{li-etal-2022-hiclre} and \textit{NLI}-based validation (DSRE + NLI) ~\cite{zhou2023improving}. These techniques are relevant forOCR-derived noisy data.  

\paragraph{Efficiency in Relation Extraction.}
Beyond accuracy, current evaluations increasingly emphasize computational efficiency. Shared tasks such as \textit{SustaiNLP 2020} measured energy consumption~\cite{wang2020sustainlp}, and \textit{EfficientQA} enforced memory limits~\cite{min2021efficientqa}. Efficiency-focused modeling continues to gain traction~\cite{boylan-etal-2025-glirel}. By combining accuracy and efficiency-focused evaluation in a historical multilingual RE task, HIPE-2026 establishes a new benchmark for robust and sustainable NLP.

\section{Conclusion}

HIPE-2026 extends the HIPE evaluation series into new research directions. 
It promotes the development of robust methods for extracting person--place relations from multilingual historical text, enabling applications in temporal and spatial analysis across the humanities and social sciences, as well as in the analysis and interpretation of literary texts. 

To formalize and isolate this challenge, the task is framed as a relation classification problem conditioned on a person--place pair within its document context. 
This formulation also facilitates the use of generative AI systems, which are expected to perform well on such contextual reasoning tasks even when applied to historical, digitized texts that diverge from clean modern conditions. 
\newline\newline
\small
\noindent\textbf{Acknowledgements.} HIPE-2026 is funded through the project \textit{Impresso – Media Monitoring of the Past II. Beyond Borders: Connecting Historical Newspapers and Radio}; SNSF 213585 and Luxembourg National Research Fund 17498891.
\newline\newline
\noindent\textbf{Disclosure of Interests.} The authors have no competing interests to declare.

\normalsize
\bibliographystyle{splncs04}
\bibliography{hipe-2026}

\end{document}